\PassOptionsToPackage{dvipsnames,table}{xcolor} 
\documentclass[journal,twoside,web]{ieeecolor}
\usepackage{jsen}
\usepackage{cite}
\usepackage{amsmath,amssymb,amsfonts}
\usepackage{bm} 
\usepackage{subcaption}
\usepackage{multirow}
\usepackage{graphicx}
\usepackage{textcomp}
\usepackage{wrapfig}
\usepackage{siunitx}
\def\BibTeX{{\rm B\kern-.05em{\sc i\kern-.025em b}\kern-.08em
    T\kern-.1667em\lower.7ex\hbox{E}\kern-.125emX}}
\definecolor{abstractbg}{rgb}{0.89804,0.94510,0.83137}
\usepackage{booktabs}

\usepackage{algorithm} 
\usepackage{algpseudocode}
\usepackage[hidelinks]{hyperref}

\newcommand{\orcidicon}[1]{%
  \href{https://orcid.org/#1}{\includegraphics[width=1em]{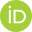}}%
}

\makeatletter
\renewcommand{\ps@titlepagestyle}{%
  \def\@oddhead{}\def\@evenhead{}%
  \def\@oddfoot{}\def\@evenfoot{}%
}
\makeatother

\begin{document}
\title{Kalman Prototypical Networks for Few-shot Fault Detection in Combined Cycle Gas Turbines}
\author{Mohammed Ayalew Belay\textsuperscript{\orcidicon{0000-0003-2144-9790}}, 
\IEEEmembership{Graduate Student Member, IEEE}, 
Lucas Ferreira Bernardino\textsuperscript{\orcidicon{0000-0002-0058-2739}},
Adil Rasheed\textsuperscript{\orcidicon{0000-0003-2690-983X}},
Rubén M. Montañés\textsuperscript{\orcidicon{0000-0002-6600-5512}},
Pierluigi Salvo Rossi\textsuperscript{\orcidicon{0000-0001-6834-8482}}, \IEEEmembership{Senior Member, IEEE}
\thanks{This work was partially supported by the Research Council of Norway under the project
DIGITAL TWIN within the PETROMAKS2 framework (project nr. 318899).}
\thanks{M.A. Belay is with the Dept. Electronic Systems, Norwegian University of Science and Technology, 7034 Trondheim, Norway (e-mail: mohammed.a.belay@ntnu.no).} 
\thanks{L. Ferreira Bernardino and R.M. Monta\~{n}es are with the Dept. Gas Technology, SINTEF Energy Research, 7491 Trondheim, Norway (e-mail: lucas.bernardino@sintef.no, ruben.mocholi.montanes@sintef.no).}
\thanks{A. Rasheed is with the Dept. Engineering Cybernetics, Norwegian University of Science and Technology, 7034 Trondheim, Norway (e-mail:  adil.rasheed@ntnu.no).}
\thanks{P. Salvo Rossi is with the Dept. Electronic Systems, Norwegian University of Science and Technology, 7034 Trondheim, Norway, and with the Dept. Gas Technology, SINTEF Energy Research, 7491 Trondheim, Norway (e-mail: salvorossi@ieee.org).}
\thanks{Manuscript received Month 00, 2025; revised Month 00, 2025.}}
\IEEEtitleabstractindextext{%
\fcolorbox{abstractbg}{abstractbg}{%
\begin{minipage}{\textwidth}%
\begin{wrapfigure}[13]{r}{3in}%
\raisebox{-10.2em}[-5em]{\includegraphics[width=2.93 in]{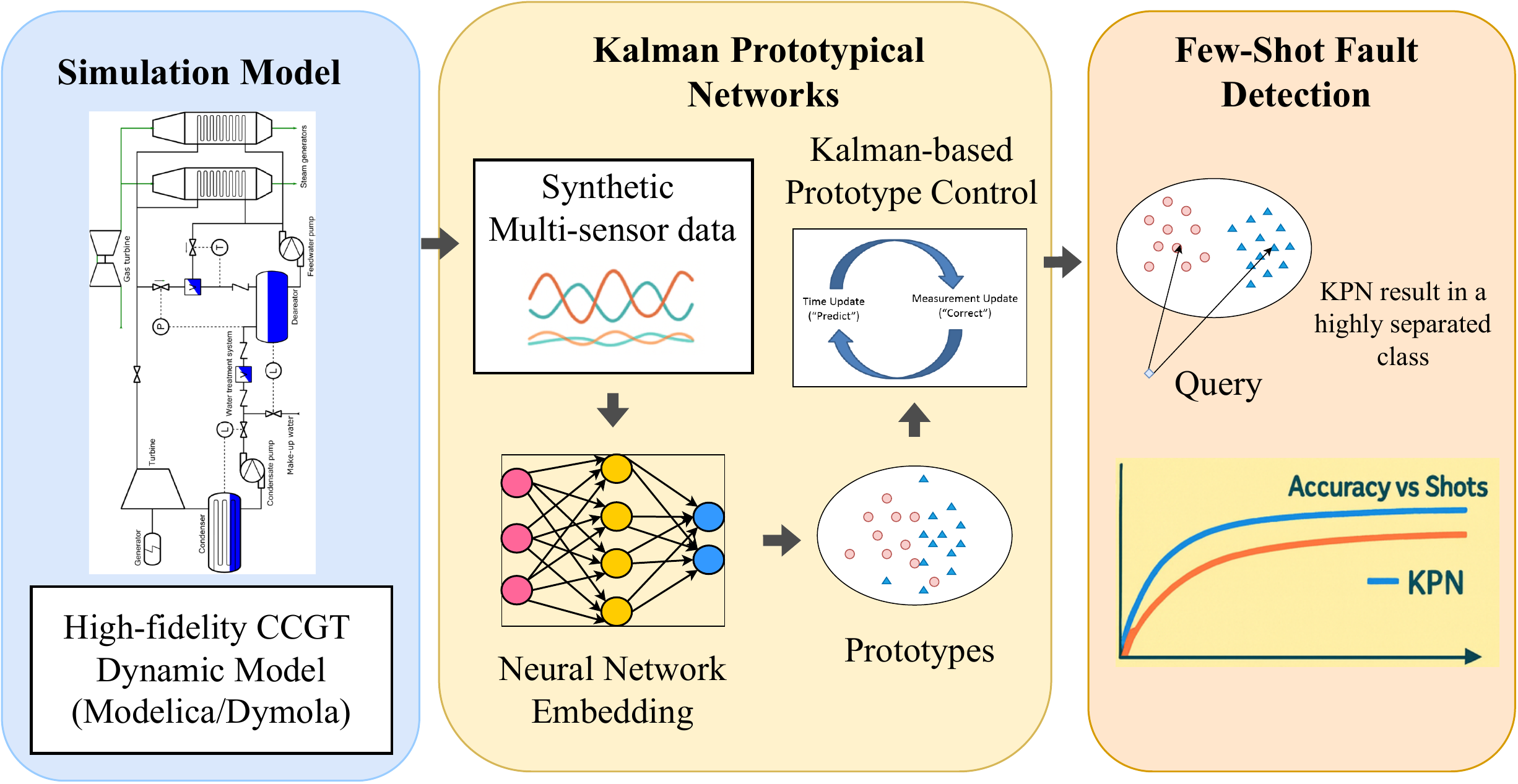}}%
\end{wrapfigure}%
\begin{abstract}
Combined-cycle gas turbines (CCGTs) play a key role in modern power generation, offering both high efficiency and reduced environmental impact. However, their complex thermo-fluid and mechanical interactions complicate fault detection, particularly when labeled fault data are scarce. In this paper, we introduce the Kalman Prototypical Network (KPN), a metric-based few-shot learning (FSL) framework specifically tailored for CCGT fault diagnosis. We model the evolution of class prototypes as latent stochastic states in a dynamic system to reduce episodic variance and improve robustness in embedding representation. Synthetic data sets generated with a high-fidelity Modelica-based dynamic simulation of an offshore CCGT system were used, simulating both normal operation and progressive leak faults under transient conditions. Application of the proposed framework on simulated leak fault detection tasks demonstrate that KPN outperforms conventional FSL methods such as Matching Networks, Relation Networks, and MAML in both accuracy and stability under varying support and query configurations. The proposed framework significantly improves training convergence and generalization by stabilizing class representations, making it well-suited for real-world CCGT fault detection where labeled data is limited.
\end{abstract}
\end{minipage}}}

\maketitle

\section{Introduction}

Combined-cycle gas turbines (CCGT) have emerged as crucial components in modern power generation, particularly favored due to their high efficiency, reliability, and environmental benefits. CCGT power plants couple a gas turbine with a heat-recovery steam generator (HRSG) and a downstream steam turbine, achieving thermal efficiencies often exceeding 60\% \cite{Gonzalez-Salazar2018ReviewRenewables}. This high efficiency has driven widespread onshore deployment and growing interest in offshore installations—where recovered excess heat can meet platform power demands while cutting CO$_2$ emissions by up to 25\% and making CCGTs favorable for modern low-emission power generation \cite{Mazzetti2021AchievingCycles}. Despite their efficiency gains, CCGT systems are inherently complex, featuring tightly coupled thermo-fluid and mechanical subsystems. Even small faults, such as a single tube leak in the HRSG, can cause high-pressure steam or water to escape, reducing net power output, damaging downstream machinery, forcing unplanned shutdowns, and introducing safety and environmental hazards. Consequently, maintaining reliability and preventing costly downtime, therefore, relies on early and accurate detection of such anomalous events.

Recently, several CCGT fault-detection methods have been proposed, including physics-based models and supervised machine-learning models. Physics-based methods rely on high-fidelity dynamic simulations to capture the full transient behavior of a CCGT under varying loads. Supervised machine learning methods utilize real-world operational labeled datasets covering each fault mode to train models \cite{Chandola2009AnomalySurvey, Belay2026Agentic, belay2026digitaltwindrivencommunicationefficientfederated}. However, CCGT fault detection faces significant challenges, primarily due to the scarcity of accurately labeled datasets. Real-world fault occurrences in CCGTs are rare, data is frequently inaccessible due to confidentiality, and manual labeling of sensor streams is both time-consuming and resource-intensive. As a result, labeled fault data remain scarce and costly to obtain, creating a critical bottleneck for deploying reliable, data-driven diagnostic systems \cite{Belay2024MTAD:Detection, tabella2025failure,Belay2023UnsupervisedDirections}. To address such challenges, fault detection methods that can learn from limited or unlabeled fault examples, such as unsupervised anomaly detection and few-shot learning (FSL), are employed in various safety-critical industrial operations.

Few-shot learning frameworks enable machine learning models to generalize from minimal labeled examples, making them ideally suited to CCGT industrial domains, where labeling costs are prohibitive and fault data are scarce. Metric-based FSL methods, such as prototypical networks, have demonstrated efficient fault detection and classification by learning robust embeddings that generalize well from very few labeled samples. These networks classify data based on proximity to class-specific prototypes computed from support sets, effectively creating representative anchors in the embedding space. However, prototypical networks suffer from significant instability due to training episodic variance—where the prototypes derived from small support sets can vary substantially between episodes, adversely affecting model robustness and classification accuracy. This episodic instability arises because each training episode may contain different subsets of data, leading to variability in the computed prototypes. Consequently, despite their promise, this instability presents a critical barrier to effective deployment in real-world industrial anomaly detection scenarios. To address this challenge, we proposed a robust Kalman prototypical network for few-shot fault detection in CCGT systems. Moreover, we utilize data generated with a high-fidelity Modelica/Dymola dynamic simulation model of CCGT to generate synthetic faults that provided the necessary data for algorithm development, benchmarking, and ultimately robust deployment of advanced diagnostic capabilities in safety-critical CCGT operations \cite{Belay2025UnsupervisedPlants}. Specifically, the primary contributions of this paper are summarized as follows:
\begin{itemize}
    \item We propose a Kalman-based Prototypical Network (KPN) that models prototype evolution as a latent stochastic state and allows for stable few-shot representation learning. 
    
   \item We utilize synthetic data sets from a high-fidelity dynamic model of an existing offshore CCGT system. 
   The data sets are extensive and represent realistic performance under transient operating conditions, simulating both normal operational dynamics and leak-induced anomalies.
    
    \item We performed an extensive performance analysis using baseline few-shot learning algorithms.
    
\end{itemize}

The remainder of this paper is structured as follows. 
Section II reviews related work and provides background on supervised and unsupervised anomaly detection methods and few-shot learning. 
Section III presents the proposed Kalman Prototypical Network method. 
Section IV describes the experimental setup, including dataset details and implementation specifics. 
Section V presents and discusses the experimental results, demonstrating the effectiveness of KPN in stabilizing prototypes and enhancing anomaly detection performance. 
Finally, Section VI concludes the paper and outlines directions for future research.

\section{Related Works}
\subsection{Fault detection in CCGT systems}

Fault detection and diagnosis (FDD) in CCGT systems are crucial for operational efficiency, safety, and reliability \cite{LIU2024IntelligentReview,Surase2024ThermalReview}. Several methods have been developed, including model-based, data-driven, and hybrid approaches \cite{Belay2025DigitalIoT, Belay2025DigitalDetection}. In particular, tube leaks are common in HRSG and OTSG units, often due to thermo‑mechanical fatigue, corrosion, or water/steam quality issues \cite{Tina2023FailureTube,Wijayanti2021FailurePlant,Patil2018ProcessModel}. Undetected leaks cause production losses and costly repairs \cite{CaseJournal}.

Model-based fault detection methods use mathematical models to represent system behavior \cite{Pourbabaee2016SensorEngines}. Data-driven methods, such as neural networks, support vector machines, and deep learning, utilize historical and real-time data to detect faults without requiring detailed physical models \cite{Camporeale2009FaultNetworks, Nayeri2022FaultComputation, Fahmi2024AdvancementsModel, Barrera2022FaultTurbines}. Hybrid approaches combine physical models with data-driven techniques \cite{Chao2019HybridModels}. Pourbabaee et al.~\cite{Pourbabaee2016SensorEngines} proposed a gas turbine sensor fault detection, isolation, and identification (FDII) method based on multiple hybrid Kalman filters (MHKFs). Camporeale et al.~\cite{Camporeale2009FaultNetworks} introduced a fault diagnosis system for CCGTs based on feed-forward neural networks. Nayeri et al.~\cite{Nayeri2022FaultComputation} proposed a Fault Detection and Isolation (FDI) system based on an ensemble-based hierarchical classifier. Ajami et al.~\cite{Ajami2012DataICA} explore independent component analysis (ICA) for fault detection and identification in the turbine system of a thermal power plant. Fahmi et al.~\cite{Fahmi2024AdvancementsModel} proposed a temporal convolutional autoencoder for gas turbine fault diagnosis using vibration data. Sarwar et al.~\cite{Sarwar2024HybridEngine} presented a multi-sensor data fusion framework for fault detection and diagnosis in an industrial gas turbine engine. Barrera et al.~\cite{Barrera2022FaultTurbines} introduce clustering and autoencoders to train predictive maintenance algorithms. Sampath et al.~\cite{Sampath2002FaultTurbine} propose a hybrid approach that combines real-world sensor data and information from simulation models. Fast et al.~\cite{Fast2010ApplicationPlant} applied artificial neural networks to monitor the condition and diagnose faults in a combined heat and power plant. Davallo et al.~\cite{Davallo2019FaultELM} proposed an Extreme Learning Machine framework for the detection and identification of leaks in an onshore CCGT. Chao et al.~\cite{Chao2019HybridModels} proposed a hybrid approach combining physical performance models with deep learning algorithms.

\subsection{Few-shot Fault detection}

In industrial settings, acquiring labeled fault data is often challenging due to the rarity of fault occurrences and the high costs associated with data annotation \cite{Zhang2022IntelligentExtensions, Li2024SmallReview}. Conventional supervised learning models, which rely heavily on extensive labeled datasets, often underperform in data-constrained conditions \cite{Belay2025SparseDetection, Belay2025AutoregressiveDetection, Belay2024MultivariateDecomposition}. Few-shot learning (FSL) has emerged as a promising solution to this problem, enabling models to generalize from a limited number of labeled examples \cite{Ren2023ADiagnosis,Liang2023Few-ShotReview}. Several specialized FSL frameworks have been developed that include metric-based, optimization-based, and memory-based approaches. Meta-based methods, such as prototypical networks, matching networks, and relation networks, focus on learning discriminative latent embedding spaces to perform classification based on similarity measures. \cite{Snell2017PrototypicalLearning, Vinyals2016MatchingLearning, Sung2018LearningLearning}. Optimization-based methods, such as Model-Agnostic Meta-Learning (MAML), learn optimization strategies to quickly adapt to new tasks using limited samples \cite{Ravi2017OptimizationLearning,Finn2017Model-agnosticNetworks}. 
Memory-based methods incorporate external memory structures or attention mechanisms to store and retrieve information efficiently \cite{Santoro2016Meta-learningNetworks}.

Recent studies have explored various FSL approaches for fault detection and diagnosis. Zhang et al. \cite{Zhang2021Few-ShotMeta-Learning} introduced a few-shot learning framework for bearing fault diagnosis based on MAML, demonstrating its superiority over traditional methods in scenarios with limited labeled data. Qiao et al. \cite{Qiao2025FaultMethod} presented a few-shot fault diagnosis model for wind turbine (WT) generators employing a Convolutional Normalization Transformer Encoder (CNTE) based on MAML. Zhang et al. \cite{Zhang2024Few-ShotPlants} introduces a prototypical network few-shot learning approach for anomaly detection in nuclear power plants. Ren et al. \cite{Ren2023Few-ShotImbalance} proposed a few-shot GAN, which uses a sample-rich class to provide a sample distribution paradigm for the sample-poor class. Zheng et al. \cite{Zheng2023Few-shotNetwork} proposed fault diagnosis based on an improved meta-relation network. Despite few-shot fault detection and diagnosis in various industrial domains, challenges remain in ensuring the stability and robustness of these models, especially under varying operational conditions. Moreover, FSL on CCGT systems remains unexplored. Addressing these issues is crucial for the reliable deployment of FSL models in real-world CCGT fault detection scenarios.

\section{The Proposed Method}

In this section, we present the proposed robust prototypical network called Kalman Prototypical Network (KPN) for few-shot fault detection in CCGT systems.

\begin{figure*}[t!]\centering
    \includegraphics[width=0.95\linewidth]{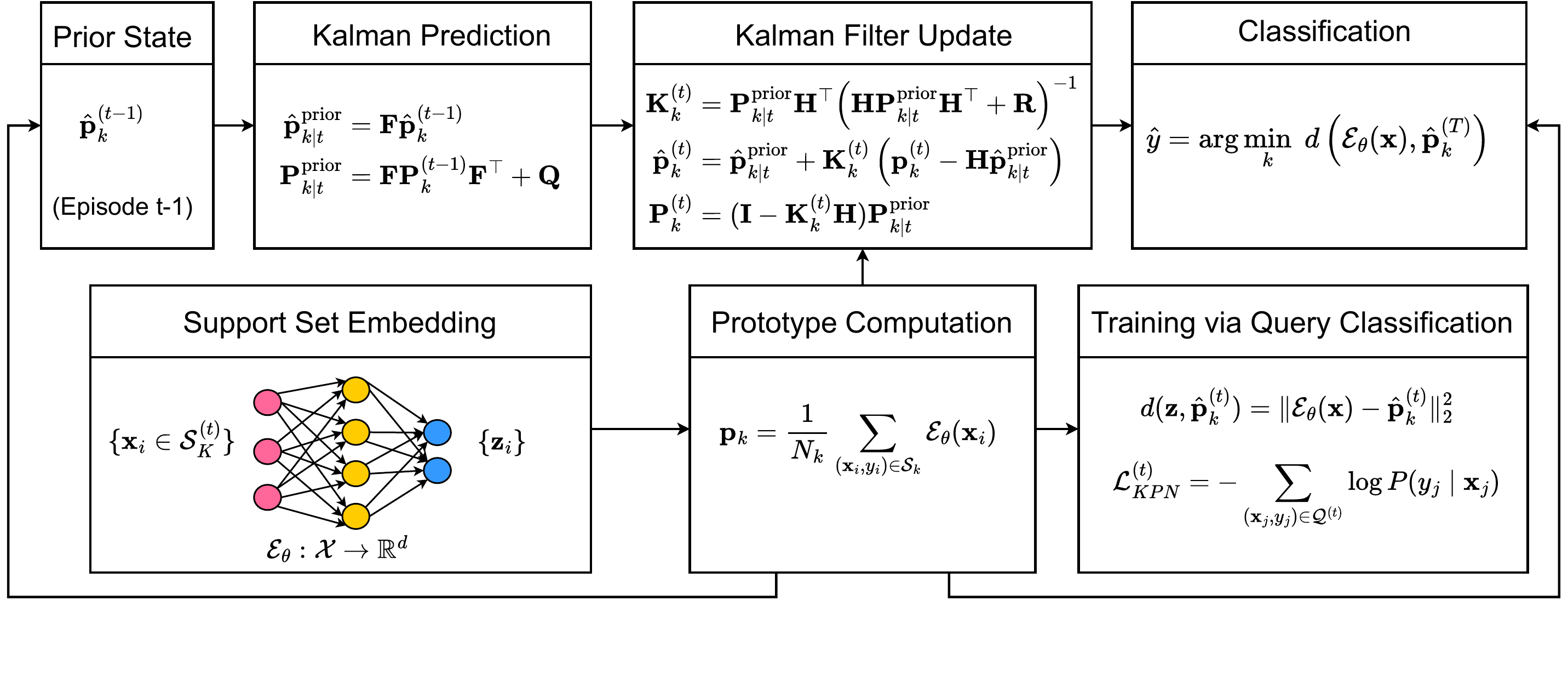}
    \caption{Kalman Prototypical Network framework (per class \(k\) per episode \(t\)).}
    \label{fig:CCGT_kalman_2}
\end{figure*}

\subsection{Prototypical Networks}

\begin{figure}[t!]\centering
    \includegraphics[width=\linewidth]{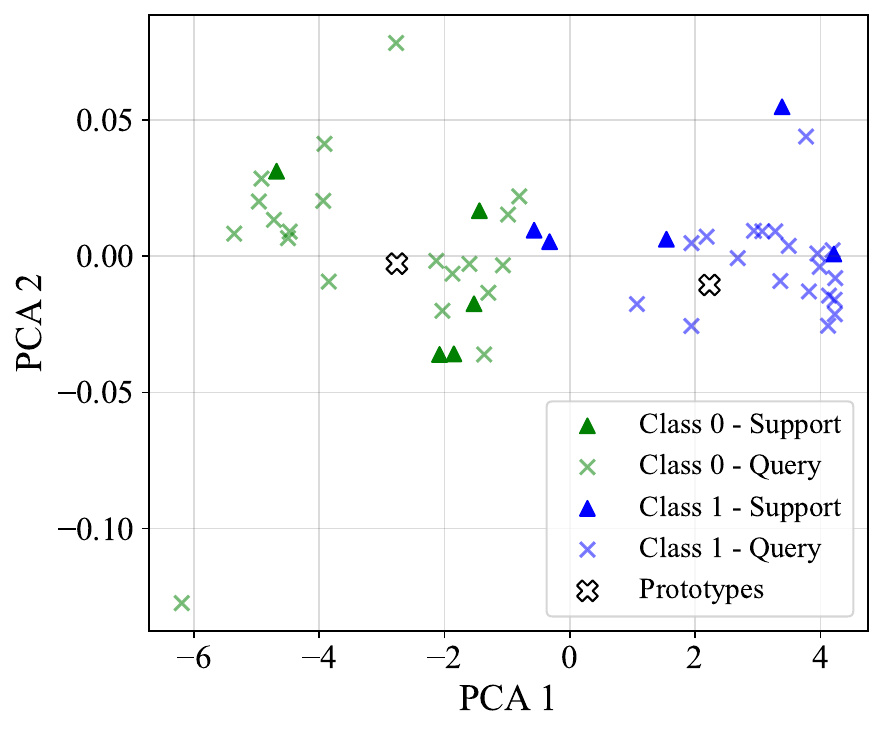}
    \caption{PCA reduced embeddings: Support, Query, Prototypes for a single episode on gas turbine dataset}
    \label{fig:embedding}
\end{figure}

Prototypical networks are a class of metric-based few-shot learning models that classify examples based on their proximity to class-specific \textit{prototypes} in a learned embedding space \cite{Snell2017PrototypicalLearning}. Let \( \mathcal{E}_\theta: \mathcal{X} \rightarrow \mathbb{R}^d \) be an embedding function parameterized by \( \theta \), which maps an input \( \mathbf{x} \in \mathcal{X} \) to a \( d \)-dimensional latent representation \( \mathcal{E}_\theta(\mathbf{x}) \in \mathbb{R}^d \). In each few-shot training \textit{episode}, the model is provided with data partitioned into a \textit{support set} \( \mathcal{S} = \bigcup_{k=1}^K \mathcal{S}_k \), where \( \mathcal{S}_k = \{ (\mathbf{x}_i, y_i) \in \mathcal{X} \times \mathcal{Y} \mid y_i = k \} \) contains \( N_k \) labeled examples of class, \( k \in \{1, \dots, K\} \)
and a disjoint \textit{query set} \( \mathcal{Q} \subset \mathcal{X}\)  for classification. Prototypical networks perform few-shot classification by computing a representative \emph{prototype} for each class using a learned embedding function. For each class \( k \), the class prototype \( \mathbf{p}_k \in \mathbb{R}^d \) is defined as the mean of embedded support examples from class \( k \):
\begin{equation}
    \mathbf{p}_k = \frac{1}{N_k} \sum_{(\mathbf{x}_i, y_i) \in \mathcal{S}_k} \mathcal{E}_\theta(\mathbf{x}_i)
\end{equation}
This forms a set of class prototypes $\mathcal{P} = \{ \mathbf{p}_1, \ldots, \mathbf{p}_K \} \subset \mathbb{R}^d$, which act as anchors for class-conditional distributions in the embedding space.  Given a query example $\mathbf{x} \in \mathcal{Q}$, we embed it using $\mathcal{E}_\theta$ and compute the squared Euclidean distance to each class prototype.
The distances are converted into class probabilities using a softmax over negative distances. This formulation is equivalent to modeling each class as a spherical Gaussian in latent space, centered at its prototype, with shared isotropic covariance.
The model is trained by minimizing the negative log-likelihood over query labels. Figure \ref{fig:embedding} depicts support, query, and prototype in the principal component analysis (PCA) reduced embedding.

\subsection{Prototype Trajectory Instability}

While prototypical networks assume that class prototypes \( \mathbf{p}_k \) are reliable and representative centroids of class embeddings within an episode, in practice, these prototypes are \textit{non-stationary} over the course of training. That is, the prototype for a given class \( k \), denoted as \( \mathbf{p}_k^{(t)} \in \mathbb{R}^d \) an episode \( t \), may vary significantly across episodes due to stochastic sampling of support sets, instability in the embedding function \( \mathcal{E}_\theta \), and shifts in the latent structure of the data.

Let \( \{ \mathbf{p}_k^{(t)} \}_{t=1}^{T} \) denote the sequence of prototypes for class \( k \) across \( T \) training episodes. Our empirical observation (e.g., via PCA or t-SNE projections) shows that:
\begin{itemize}
    \item Prototypes often follow a smooth but nonlinear trajectory through the latent space.
    \item The distance \( \| \mathbf{p}_k^{(t+1)} - \mathbf{p}_k^{(t)} \| \) is non-negligible, indicating episoidal drift.
    \item Prototypes may oscillate or diverge, particularly in high-variance regimes or early training stages.
\end{itemize}
Such behavior suggests that the episodic average used in vanilla prototypical networks may be an inconsistent estimate of the true latent class centroid, particularly when support sets are small or non-representative.
We reinterpret prototype evolution as a time series \( \{ \mathbf{p}_k^{(t)} \}_{t=1}^{T} \), generated by a latent dynamic process. The goal is to recover a \textit{denoised} \( \hat{\mathbf{p}}_k^{(t)} \) of the prototype at each episode. We utilize Kalman filters to reduce \textit{prototype variance} ($\sigma_k^2$), which use both current observations and prior history to optimally estimate latent variables in dynamic systems. 
Higher \( \sigma_k^2 \) is often associated with degraded generalization due to inconsistent decision boundaries for query samples. By enforcing smoothing across episodes, we aim to reduce \( \sigma_k^2 \), improve \textit{stability}, and ultimately enhance \textit{robustness during test-time classification}.

\subsection{Kalman-Based Prototype Estimation}

To address the instability in the prototype sequence \( \{ \mathbf{p}_k^{(t)} \}_{t=1}^T \), we propose modeling prototype evolution as a latent linear dynamical system. Specifically, we treat the true class prototype \( \hat{\mathbf{p}}_k^{(t)} \in \mathbb{R}^d \) at episode \( t \) as a latent state, and the observed prototype \( \mathbf{p}_k^{(t)} \) as a noisy measurement of that state.


\subsubsection*{Prototype State Space Model}

We assume the following discrete-time linear Gaussian system for each class \( k \):

\begin{itemize}
    \item{State transition equation (process model):}
    \begin{equation}
            \hat{\mathbf{p}}_k^{(t)} = \mathbf{F} \hat{\mathbf{p}}_k^{(t-1)} + \mathbf{w}_k^{(t)}, \quad \mathbf{w}_k^{(t)} \sim \mathcal{N}(\mathbf{0}, \mathbf{Q})
    \end{equation}
   
    \item {Observation equation (measurement model):}
        \begin{equation}
    \mathbf{p}_k^{(t)} = \mathbf{H} \hat{\mathbf{p}}_k^{(t)} + \mathbf{v}_k^{(t)}, \quad \mathbf{v}_k^{(t)} \sim \mathcal{N}(\mathbf{0}, \mathbf{R})
    \end{equation}
\end{itemize}

Here, \( \hat{\mathbf{p}}_k^{(t)} \) is the latent prototype, \( \mathbf{p}_k^{(t)} \) is the observed prototype from the support set, and \( \mathbf{F}, \mathbf{H} \in \mathbb{R}^{d \times d} \) are the transition and observation matrices, typically set to identity. The covariance matrices \( \mathbf{Q}, \mathbf{R} \in \mathbb{R}^{d \times d} \) control the process and observation noise, respectively.

\subsubsection*{Prototype Recursive Kalman Update}

Given the prior estimate \( \hat{\mathbf{p}}_k^{(t-1)} \) with covariance \( \mathbf{P}_k^{(t-1)} \), the Kalman filter proceeds in two steps:
\paragraph{Prediction Step}
\begin{align} 
\hat{\mathbf{p}}_{k|t}^{\text{prior}} &=  \mathbf{F} \hat{\mathbf{p}}_k^{(t-1)} \\ 
\mathbf{P}_{k|t}^{\text{prior}} &=  \mathbf{F} \mathbf{P}_k^{(t-1)} \mathbf{F}^\top + \mathbf{Q}
\end{align}

\paragraph{Update Step}
\begin{align} 
\mathbf{K}_k^{(t)} &=  \mathbf{P}_{k|t}^{\text{prior}} \mathbf{H}^\top \left( \mathbf{H} \mathbf{P}_{k|t}^{\text{prior}} \mathbf{H}^\top + \mathbf{R} \right)^{-1} \\ 
\hat{\mathbf{p}}_k^{(t)} &=  \hat{\mathbf{p}}_{k|t}^{\text{prior}} + \mathbf{K}_k^{(t)} \left( \mathbf{p}_k^{(t)} - \mathbf{H} \hat{\mathbf{p}}_{k|t}^{\text{prior}} \right)\\
\mathbf{P}_k^{(t)} &=  (\mathbf{I} - \mathbf{K}_k^{(t)} \mathbf{H}) \mathbf{P}_{k|t}^{\text{prior}}
\end{align}
For prototype tracking, we assumed:
$\mathbf{F} = \mathbf{H} = \mathbf{I}, \quad \mathbf{Q} = q \cdot \mathbf{I}, \quad \mathbf{R} = r \cdot \mathbf{I}$, where \( q, r > 0 \) are tunable scalars. This yields a simplified updates:
\begin{equation}
\mathbf{K}_k^{(t)} = \frac{\mathbf{P}_{k|t}^{\text{prior}}}{\mathbf{P}_{k|t}^{\text{prior}} + r \cdot \mathbf{I}}, \quad
\hat{\mathbf{p}}_k^{(t)} = \hat{\mathbf{p}}_{k|t}^{\text{prior}} + \mathbf{K}_k^{(t)} \left( \mathbf{p}_k^{(t)} - \hat{\mathbf{p}}_{k|t}^{\text{prior}} \right)
\end{equation}

During episodic training, we maintain a filtered prototype \( \hat{\mathbf{p}}_k^{(t)} \) for each class, updated using the above equations. These filtered prototypes are then used in place of raw episodic means when classifying query samples. This approach leads to a smoothed and denoised prototype trajectories, reduced variance in decision boundaries, and improved generalization in few-shot learning tasks.

\subsection{Kalman Prototypical Network}
We incorporated the Kalman filter into the prototypical network learning pipeline, resulting in the \textit{Kalman Prototypical Network (KPN)}. 
In our framework, the Kalman filter operates over training episodes, i.e., the time index $t$ denotes the episode count rather than physical time. The observation \( \hat{\mathbf{p}}_k^{(t)} \) is the per-episode prototype computed from a stochastically sampled support set, and the filter provides temporal smoothing across episodes to reduce episodic variance in class centroids. This differs from real-time or online training, where $t$ would index sequential measurements and the filter would update as new data arrived in chronological order. 
The central idea is to treat each class prototype \( \hat{\mathbf{p}}_k^{(t)} \in \mathbb{R}^d \) as a latent state variable, which evolves over training episodes \( t = 1, \ldots, T \). The observed prototypes \( \mathbf{p}_k^{(t)} \), computed from episodic support sets, are viewed as noisy measurements of these latent states.

For a given query example \( \mathbf{x} \in \mathcal{Q}^{(t)} \), with embedding \(\mathbf{z}=\mathcal{E}_\theta(\mathbf{x}) \), classification is  based on the squared Euclidean distance to the \textit{Kalman-filtered prototypes}:
\begin{equation}
   d(\mathbf{z},\hat{\mathbf{p}}_k^{(t)}) = \| \mathcal{E}_\theta(\mathbf{x}) - \hat{\mathbf{p}}_k^{(t)} \|_2^2 = (\mathcal{E}_\theta(\mathbf{x}) - \hat{\mathbf{p}}_k^{(t)})^\top(\mathcal{E}_\theta(\mathbf{x}) - \hat{\mathbf{p}}_k^{(t)}) 
\end{equation}

The distances are converted into class probabilities using a
softmax over negative distances:
\begin{equation}
    P(y = k \mid \mathbf{x}) = \frac{\exp\left( -\| \mathcal{E}_\theta(\mathbf{x}) - \hat{\mathbf{p}}_k^{(t)} \|^2 \right)}{\sum_{k'=1}^K \exp\left( -\| \mathcal{E}_\theta(\mathbf{x}) - \hat{\mathbf{p}}_{k'}^{(t)} \|^2 \right)}.
\end{equation}

The model is trained by minimizing the negative log-likelihood over query labels.
The total loss for the episode \( t \)  is given by:
\begin{equation}
    \mathcal{L}_{KPN}^{(t)} = - \sum_{(\mathbf{x}_j, y_j) \in \mathcal{Q}^{(t)}} \log P(y_j \mid \mathbf{x}_j).
\end{equation}


Gradient-based optimization is applied to update \( \theta \), the parameters of the embedding function, while the Kalman parameters \( \mathbf{Q}, \mathbf{R} \) are either fixed or tuned separately. 
The gradient of loss pushes the embedding $\mathcal{E}_\theta$ closer to its true prototype $\mathbf{p}_y$ while simultaneously repelling it from the other prototypes, weighted by their exponentially scaled distances.
During inference, we evaluate the model on query examples \( \mathcal{Q}^{\text{test}} \) using a fixed prototype \( \hat{\mathbf{p}}_k^{(*)} \) for each class \( k \). We use the last filtered prototype (i.e. \( \hat{\mathbf{p}}_k^{(*)} = \hat{\mathbf{p}}_k^{(T)} \)) for inference. The model predicts the label \( \hat{y} \in \{1, \dots, K\} \) by selecting the minimum distance from all class prototypes.
\begin{equation}
\hat{y} = \arg\min_k \; d\left( \mathcal{E}_\theta(\mathbf{x}), \hat{\mathbf{p}}_k^{(T)} \right)
\end{equation}
This can be interpreted as a linear classifier in the embedding space when the prototypes are fixed and distances are expanded as inner products. The training algorithm for KPN is summarized in Algorithm \ref{alg:kpn_training}.

\algnewcommand\algorithmicinput{\textbf{Input:}}
\algnewcommand\algorithmicoutput{\textbf{Output:}}
\algnewcommand\Input{\item[\algorithmicinput]}%
\algnewcommand\Output{\item[\algorithmicoutput]}%
\begin{algorithm}[t!]
\caption{Kalman Prototypical Network (Training)}
\label{alg:kpn_training}
\begin{algorithmic}[1]
\Input Episodes $\{E_t\}_{t=1}^T$, where $E_t=(S^{(t)}, Q^{(t)})$; classes $\mathcal{K}=\{1,\dots,K\}$; embedding $E_\theta(\cdot)$; Kalman parameters $F=I$, $H=I$, $Q=qI$, $R=rI$; initialization $\hat p^{(0)}_{k} \gets 0$, $P^{(0)}_{k} \gets \alpha I$ for all $k \in \mathcal{K}$.
\Output Trained parameters $\theta$ and $\{\hat p^{(T)}_{k}, P^{(T)}_{k}\}_{k=1}^K$.
\For{$t = 1$ \textbf{to} $T$}
    \ForAll{$k \in \mathcal{K}$}\Comment{Episodic prototypes}
        \State $Z^{(t)}_{k} \gets \{\, E_\theta(x)\;\mid\;(x,y{=}k)\in S^{(t)} \,\}$
        \State $p^{(t)}_{k} \gets \dfrac{1}{|Z^{(t)}_{k}|}\sum_{z\in Z^{(t)}_{k}} z$
    \EndFor
    \ForAll{$k \in \mathcal{K}$}\Comment{Kalman prediction}
        \State $\hat p^{\mathrm{prior}}_{k|t} \gets \hat p^{(t-1)}_{k}$
        \State $P^{\mathrm{prior}}_{k|t} \gets P^{(t-1)}_{k} + Q$
    \EndFor
    \ForAll{$k \in \mathcal{K}$}\Comment{Kalman update}
        \State $K^{(t)}_{k} \gets P^{\mathrm{prior}}_{k|t}\!\left(P^{\mathrm{prior}}_{k|t}+R\right)^{-1}$
        \State $\hat p^{(t)}_{k} \gets \hat p^{\mathrm{prior}}_{k|t} + K^{(t)}_{k}\!\left(p^{(t)}_{k}-\hat p^{\mathrm{prior}}_{k|t}\right)$
        \State $P^{(t)}_{k} \gets \left(I-K^{(t)}_{k}\right) P^{\mathrm{prior}}_{k|t}$
    \EndFor
    \State $L^{(t)}_{\mathrm{KPN}} \gets 0$\Comment{Query classification and loss}
    \ForAll{$(x_j,y_j)\in Q^{(t)}$}
        \State $z_j \gets E_\theta(x_j)$
        \ForAll{$k \in \mathcal{K}$}
            \State $d_k \gets \|z_j - \hat p^{(t)}_{k}\|_2^2$
        \EndFor
        \State $P(y{=}k\mid x_j) \gets \mathrm{softmax}_k(-d_k)$
        \State $L^{(t)}_{\mathrm{KPN}} \gets L^{(t)}_{\mathrm{KPN}} - \log P(y{=}y_j\mid x_j)$
    \EndFor
    \State $\theta \gets \mathrm{OptimizerStep}\!\left(\theta, \nabla_\theta L^{(t)}_{\mathrm{KPN}}\right)$\Comment{Update embedding parameters}
\EndFor
\end{algorithmic}
\end{algorithm}

\subsection{Computational Complexity}

The Kalman Prototypical Network improves the standard prototypical framework by providing statistically optimal estimates under Gaussian noise assumptions. This comes with an additional cost that arises solely from the recursive update of KPN prototypes per episode and per class. Let $d$ denote the embedding dimensionality, $K$ the number of classes per episode, and $T$ the total number of training episodes.
Each update requires matrix addition and inversion in $\mathbb{R}^{d \times d}$, resulting in complexity $\mathcal{O}(d^3)$ per class per episode. The total overhead for all $K$ classes is $\mathcal{O}(K d^3)$ per episode, which is negligible for moderate $d$. In comparison, standard Prototypical Networks involve computing class means from $N$ support samples with complexity $\mathcal{O}(N d)$ and query-class distances for $M$ queries with complexity $\mathcal{O}(M K d)$
Thus, the dominant computational cost remains in embedding and distance calculations, and the added filtering step introduces minimal complexity.
The episodic filtered prototype variance is given by 
\begin{equation}
\tilde{\sigma}_k^2 = \frac{1}{T} \sum_{t=1}^T \left\| \hat{\mathbf{p}}_k^{(t)} - \tilde{\mathbf{p}}_k \right\|_2^2, \quad
\tilde{\mathbf{p}}_k = \frac{1}{T} \sum_{t=1}^T \hat{\mathbf{p}}_k^{(t)}
\end{equation}
In general, the Kalman filter reduces the variance: 
$\tilde{\sigma}_k^2 \leq \sigma_k^2, \; \forall k$.
This reduction improves temporal smoothness and inter-episode consistency of class representations, which in turn enhances generalization performance. The Kalman filter yields the maximum a posteriori estimate:
\begin{equation}
\hat{\mathbf{p}}_k^{(t)} = \arg\min_{\hat{\mathbf{p}}} \left\| \hat{\mathbf{p}} - \mathbf{p}_k^{(t)} \right\|_{\mathbf{R}^{-1}}^2 + \left\| \hat{\mathbf{p}} - \hat{\mathbf{p}}_k^{(t-1)} \right\|_{\mathbf{Q}^{-1}}^2
\end{equation}
thus, the Kalman filter in KPN acts as a Bayesian smoother, optimally denoising the prototype under Gaussian assumptions.


\section{Experimental Setup}
\begin{figure*}[ht!]\centering
\includegraphics[width=0.75\linewidth]{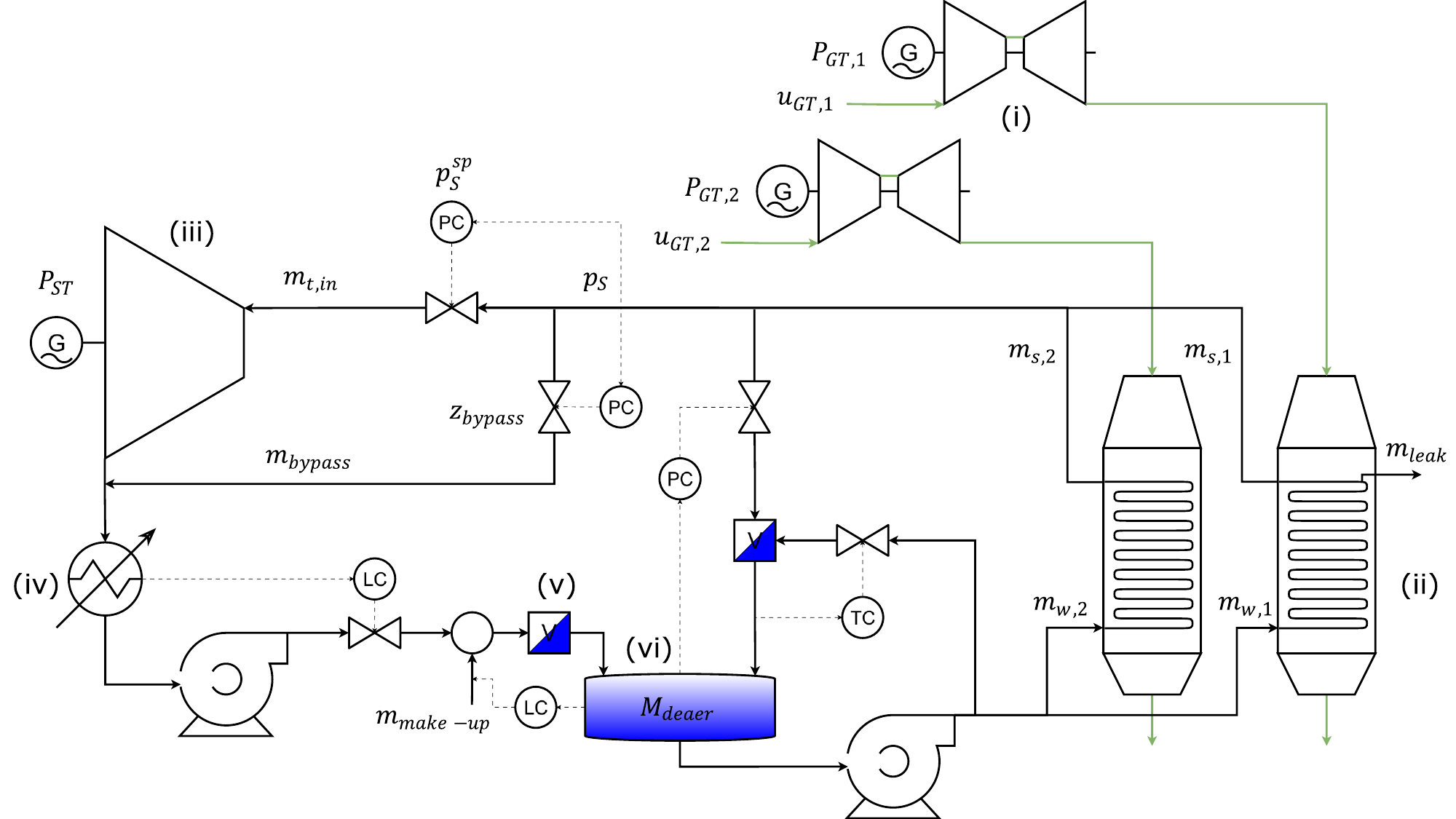}
\caption{Schematic of the steam cycle as implemented in the dynamic model \cite{Belay2025UnsupervisedPlants,Montanes2023AEnvironments}.}  
\label{fig:method}
\end{figure*}

\subsection{Dataset}

In a previous work by the authors \cite{Belay2025UnsupervisedPlants}, we generated five time series datasets corresponding to the development of a leak at the OTSG steam header, utilizing the dynamic simulation model \cite{Montanes2023AEnvironments}. A schematic of the simulated CCGT system is shown in Figure~\ref{fig:method}. The main input and output variables of the dynamic model relevant for this work are summarized in Tables~\ref{tab:input} and \ref{tab:output}. For each time series, the steam cycle and OTSGs are subject to operation under the normal variability of the connected GTs, with load oscillations around the nominal point and step changes corresponding to different operating nominal loads. The normal variability is based on historical operational data for mechanical drive GTs during a year that were previously analyzed for a reference offshore platform in \cite{Montanes2021CompactGenerators}. Each of the five time series has a duration of 1 week, with a sampling resolution of 1 hour. Three of the time series correspond to normal operation (without OTSG leaks), and the remaining two time series correspond to the gradual development of a leak in one of the OTSGs. Faulty data was simulated by increasing the size of the leak over time (orifice opening), resulting in an increasing mass flow rate of water/steam through the leak over time.

\begin{table}[ht]\centering
\caption{Input parameters for dataset generation \cite{Belay2025UnsupervisedPlants}.}
\renewcommand{\arraystretch}{1.0} 
\resizebox{1.\columnwidth}{!}{%
\begin{tabular}{lll}
\toprule
\textbf{Variable} & \textbf{Description} & \textbf{Nominal value} \\ \midrule
$u_{GT,1}$          & Load input for GT 1                       & 80.76 \%                \\
$u_{GT,2}$          & Load input for GT 2                       & 80.76 \%                \\
$T_{s,1}^{sp}$      & Outlet temperature set-point of OTSG 1    & 711.45 K                \\
$T_{s,2}^{sp}$      & Outlet temperature set-point of OTSG 2    & 711.45 K                \\
$p_s^{sp}$          & Steam pressure set-point                  & 1.65 $\times 10^6$ Pa   \\
\textit{valve leak} & Leak simulation input                     & 0.0                     \\
\bottomrule
\end{tabular}
}
\label{tab:input}
\end{table}

\begin{table}[ht]\centering
\caption{Output parameters from dataset generation \cite{Belay2025UnsupervisedPlants}.}
\renewcommand{\arraystretch}{1.0} 
\resizebox{0.9\columnwidth}{!}{%
\begin{tabular}{ll}
\toprule
\textbf{Variable} & \textbf{Description} \\ \midrule
$P_{GT,1}$          & Shaft power production of GT 1 (MW)             \\  
$P_{GT,2}$          & Shaft power production of GT 2 (MW)             \\  
$P_{ST}$            & Power production of ST (MW)               \\  
$T_{s,1}$           & Outlet temperature of OTSG 1 (K)          \\  
$T_{s,2}$           & Outlet temperature of OTSG 2 (K)          \\  
$m_{w,1}$           & Inlet water flowrate of OTSG 1 (kg/s)     \\  
$m_{w,2}$           & Inlet water flowrate of OTSG 2 (kg/s)     \\  
$m_{s,1}$           & Outlet steam flowrate of OTSG 1 (kg/s)    \\  
$m_{s,2}$           & Outlet steam flowrate of OTSG 2 (kg/s)    \\  
$p_s$               & Steam pressure (Pa)                       \\  
$M_{deaer}$         & Deaerator hold-up (kg)                    \\  
$m_{make-up}$       & Make-up water flowrate (kg/s)             \\  
$m_{t,in}$          & Turbine inlet steam flowrate (kg/s)       \\  
$z_{bypass}$        & Turbine bypass valve opening (-)          \\  
$m_{bypass}$        & Turbine bypass steam flowrate (kg/s)      \\  
$m_{leak}$          & Leak flowrate (kg/s)                      \\  
\bottomrule
\end{tabular}
}
\label{tab:output}
\end{table}

\subsection{Baseline Methods}
To evaluate the performance of the proposed method, we considered five state-of-the-art few-shot learning approaches.

\begin{itemize}
\item \textbf{Prototypical Network} \cite{Snell2017PrototypicalLearning} employs a neural network-based encoder and classifies queries by negative Euclidean distance to per-class prototypes.
\item \textbf{Matching Network} \cite{Vinyals2016MatchingLearning} uses the same encoder but computes cosine‐normalized embeddings and attends over support examples via softmax similarity.
\item \textbf{Relation Network} \cite{Sung2018LearningLearning} concatenates each query–support pair of embeddings and passes them through a relation module to predict similarity scores.
\item \textbf{MAML} \cite{Finn2017Model-agnosticNetworks} meta‐trains an MLP classifier by repeatedly sampling tasks, performing inner-loop SGD updates on support sets, and outer-loop meta-updates. 
\end{itemize}

\subsection{Implementation Details and Tools}

The combined dataset is split into an 80\% training set and a 20\% test set, stratified by the binary label. All features are standardized to zero mean and unit variance using a scaler fitted on the training data. For the neural network–based encoders (Prototypical, Matching, Relation), we use one hidden layer of dimension 8 and an embedding layer of dimension 4. The Relation Network’s relation module is a two‐layer MLP with hidden size 16. For MAML, we perform 30 inner‐loop gradient steps with an inner learning rate of 0.1 and a single update with an outer learning rate of 0.01.
For the proposed method (KPN), we set process noise \(q=10^{-3}\) and observation noise \(r=10^{-2}\). All models are trained for 50 episodes using the Adam optimizer. All models are trained for 50 episodes using the Adam optimizer. Models are implemented using PyTorch and trained on NVIDIA RTX A5000 GPUs.

\section{Results and Discussions}

\subsection{Fault Detection Performance}
\subsubsection{Few-shot Performance}

\begin{figure}[t!]\centering
    \includegraphics[width=\linewidth]{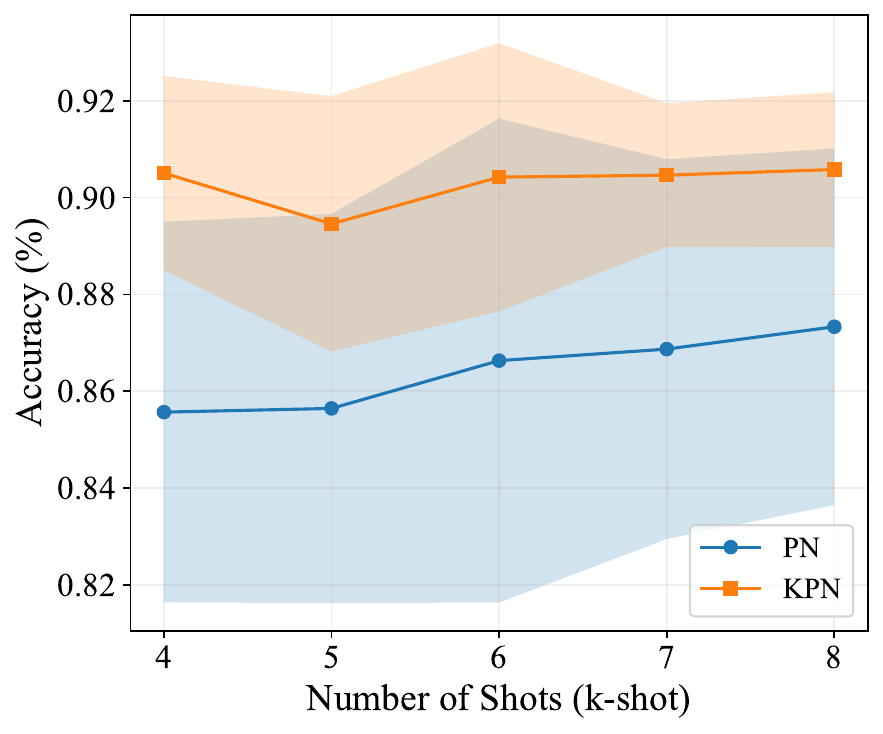}
    \caption{Accuracy vs. query per class for fixed test episodes and number of shots.}
    \label{fig:shot_vs_test_accuracy_band}
\end{figure}

We evaluate the proposed Kalman Prototypical Network (KPN) with four established few-shot learning baselines—Prototypical Network, Matching Network, Relation Network, and MAML (via Reptile optimization)—on the gas turbine fault detection task under a 2-way classification setting. As shown in Table \ref{tab:accuracy_vs_few_shot}, models are evaluated across a range of support set sizes (4-shot to 8-shot) using 100 test episodes and 20 random experiments. Reported results are the mean accuracy and standard deviation over all runs. Across all shot counts, KPN consistently achieves the highest accuracy with the lowest variance, outperforming all baselines. In the 4-shot setting, KPN achieves 90.51\% ± 2.01\%, which is approximately 1.3\% higher than MatchingNet and nearly 5\% higher than ProtoNet. The performance gains are particularly evident in lower-shot regimes, where baselines exhibit high sensitivity to prototype variance and small-sample noise. As shown in figure \ref{fig:shot_vs_test_accuracy_band}, the prototypical network suffers from relatively high variance (e.g., {±5.00\%} in 6-shot), attributed to unstable prototype estimation across episodes. Matching Network improves robustness via attention over support embeddings but remains susceptible to support-query mismatch. RelationNet performs significantly worse than other methods due to the challenges of learning a parametric relation module from limited support data, exhibiting both low accuracy and high variance. MAML demonstrates moderate accuracy but lacks stability, likely due to the difficulty of optimizing task-specific weights with very few support examples. By contrast, KPN stabilizes class representations through temporal smoothing of prototypes, yielding semantically consistent latent centroids across training episodes. This leads to superior generalization and robustness in few-shot inference, particularly under challenging low-data conditions.


\begin{table*}[ht]
\centering
\small
\setlength{\tabcolsep}{6pt}            
\renewcommand{\arraystretch}{1.5}       
\caption{Accuracy vs. number of support set for a fixed test episodes and query per class.}
\resizebox{0.9\textwidth}{!}{%
  \begin{tabular}{cccccc}
    \hline
    \multirow{2}{*}{\centering\textbf{Models}} 
      & \multicolumn{5}{c}{\textbf{Number of support set (few-shot size)}} \\ 
    \cline{2-6}
      & \textbf{4} & \textbf{5} & \textbf{6} & \textbf{7} & \textbf{8} \\
    \hline
ProtoNet            & 85.57\% $\pm$ 3.93\%   & 85.64\% $\pm$ 4.02\%   & 86.63\% $\pm$ 5.00\%   & 86.87\% $\pm$ 3.93\%   & 87.33\% $\pm$ 3.69\%   \\
MatchingNet         & 89.16\% $\pm$ 3.57\%   & 88.56\% $\pm$ 4.21\%   & 87.65\% $\pm$ 4.29\%   & 88.46\% $\pm$ 4.67\%   & 88.27\% $\pm$ 5.46\%   \\
RelationNet         & 63.88\% $\pm$ 12.52\%  & 61.06\% $\pm$ 11.55\%  & 63.48\% $\pm$ 14.09\%  & 66.17\% $\pm$ 13.03\%  & 71.25\% $\pm$ 11.85\%  \\
MAML        & 77.66\% $\pm$ 9.73\%   & 73.34\% $\pm$ 10.87\%  & 79.47\% $\pm$ 10.89\%  & 74.62\% $\pm$ 11.49\%  & 76.58\% $\pm$ 12.63\%  \\
\textbf{KPN (ours)}           & \textbf{90.51\% $\pm$ 2.01\%}   & \textbf{89.46\% $\pm$ 2.64\%}   & \textbf{90.42\% $\pm$ 2.77\%}   & \textbf{90.46\% $\pm$ 1.49\%}   & \textbf{90.58\% $\pm$ 1.60\%}   \\ \hline
  \end{tabular}%
}
\label{tab:accuracy_vs_few_shot}
\end{table*}

\subsubsection{Query size Performance}

We further evaluate the models for a fixed few-shot size, and the number of query samples per class is varied from 5 to 25. Table \ref{tab:accuracy_vs_query} presents the classification performance of four baselines and the proposed method. All models are evaluated under a fixed 5-shot configuration, using 100 test episodes and 20 random seeds. 
Across all query sizes, KPN consistently outperforms the baselines, achieving accuracies above 90\% with significantly lower variance. Notably, KPN achieves its best performance at 15 queries per class, with 90.71\% $\pm$ 2.75\%, and maintains strong stability across other settings. The minimal degradation in performance as the query size increases demonstrates KPN’s robustness and reliability under varying evaluation loads. ProtoNet displays moderate improvement with increasing query size, but its performance remains 3–5\% below that of KPN. This can be attributed to its reliance on per-episode prototypes, which introduces variability and reduces consistency in prediction. MatchingNet performs better than ProtoNet, especially at lower query counts, due to its attention-based architecture, yet still falls short of KPN in both accuracy and variance. RelationNet performs the worst across all query sizes, with high variability and accuracies ranging from 61.06\% to 71.71\%. Its learned similarity module likely struggles to generalize from few-shot support examples. MAML demonstrates moderate performance but suffers from high variance, a consequence of instability in meta-learned adaptation when limited data is available per task. 
The low standard deviation across all settings indicates that KPN achieves both high accuracy and robustness. Such stability is essential for industrial diagnostic systems, such as gas turbine fault detection, where batch size during deployment may depend on operational constraints or data streaming conditions.

\begin{figure}[t!]\centering
    \includegraphics[width=\linewidth]{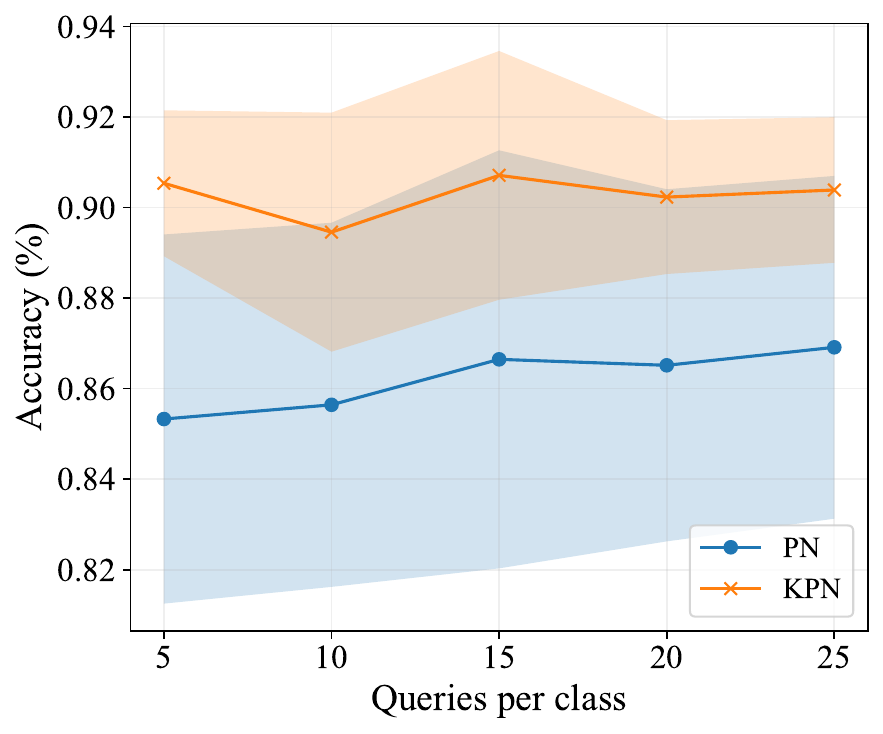}
    \caption{Accuracy vs. query per class for fixed test episodes and number of shots.}
    \label{fig:query_vs_test_accuracy_band}
\end{figure}



\begin{table*}[!ht]
\centering
\small
\setlength{\tabcolsep}{6pt}            
\renewcommand{\arraystretch}{1.5}       
\caption{Accuracy vs. query per class set for a fixed test episodes and few-shot samples}
\resizebox{0.9\textwidth}{!}{%
  \begin{tabular}{cccccc}
    \hline
    \multirow{2}{*}{\centering\textbf{Model}} 
      & \multicolumn{5}{c}{\textbf{Query-size}} \\ 
    \cline{2-6}
      & \textbf{5}        & \textbf{10}       & \textbf{15}       & \textbf{20}       & \textbf{25} \\
    \hline
ProtoNet            & 85.33\% $\pm$ 4.08\%   & 85.64\% $\pm$ 4.02\%   & 86.65\% $\pm$ 4.62\%   & 86.52\% $\pm$ 3.89\%   & 86.91\% $\pm$ 3.79\%   \\
MatchingNet         & 89.17\% $\pm$ 3.44\%   & 88.56\% $\pm$ 4.21\%   & 87.67\% $\pm$ 4.29\%   & 88.47\% $\pm$ 4.65\%   & 88.25\% $\pm$ 5.25\%   \\
RelationNet         & 62.97\% $\pm$ 12.34\%  & 61.06\% $\pm$ 11.55\%  & 63.22\% $\pm$ 14.05\%  & 66.54\% $\pm$ 12.96\%  & 71.71\% $\pm$ 11.00\%  \\
MAML        & 78.12\% $\pm$ 9.64\%   & 73.34\% $\pm$ 10.87\%  & 79.14\% $\pm$ 11.01\%  & 73.57\% $\pm$ 11.04\%  & 75.64\% $\pm$ 12.97\%  \\
\textbf{KPN (ours)}  & \textbf{90.54\% $\pm$ 1.61\%}   & \textbf{89.46\% $\pm$ 2.64\%}   & \textbf{90.71\% $\pm$ 2.75\%}   & \textbf{90.23\% $\pm$ 1.70\%}   & \textbf{90.39\% $\pm$ 1.61\%}   \\ 
 \hline
\end{tabular}
}
\label{tab:accuracy_vs_query}

\end{table*}

\subsection{Parameter Sensitivity Analysis}

\subsubsection{Process Noise and Measurement Noise} 
\begin{figure}[t!]\centering
    \includegraphics[width=\linewidth]{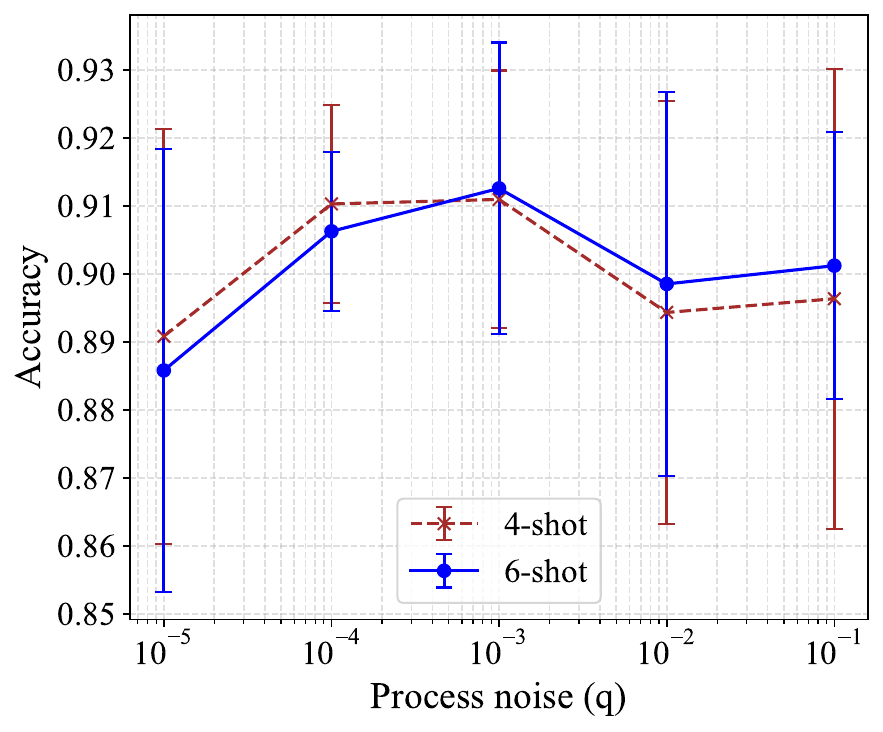}
    \caption{Accuracy vs. process noise for fixed test episodes and query per class.}
    \label{fig:process_noise}
\end{figure}

We analyze the classification accuracy of KPN under varying levels of process noise \( q \in \{10^{-5}, 10^{-4}, 10^{-3}, 10^{-2}, 10^{-1}\} \), with a fixed measurement noise \( r = 10^{-3} \), for both 4-shot and 6-shot configurations. Each data point represents the mean accuracy across 100 evaluation episodes and 20 different random experiments. As shown in Figure~\ref{fig:process_noise}, KPN achieves peak performance at a moderate process noise level, specifically \( q = 10^{-3} \), attaining an accuracy of 91.09\% (±1.90\%) for the 4-shot case and 91.25\% (±2.15\%) for the 6-shot case. This indicates that an intermediate value of \( q \) effectively balances the trade-off between adaptability and temporal smoothing of class prototypes. At very low process noise values (e.g., \( q = 10^{-5} \)), the filter becomes overly conservative, causing the prototypes to remain rigid and less responsive to changes, leading to reduced accuracy: 89.08\% (4-shot) and 88.58\% (6-shot). Conversely, excessively high process noise (e.g., \( q = 10^{-1} \)) leads to an overly reactive filter, which diminishes its smoothing capability and results in noisier prototypes and degraded performance.

\begin{figure}[t!]\centering
    \includegraphics[width=\linewidth]{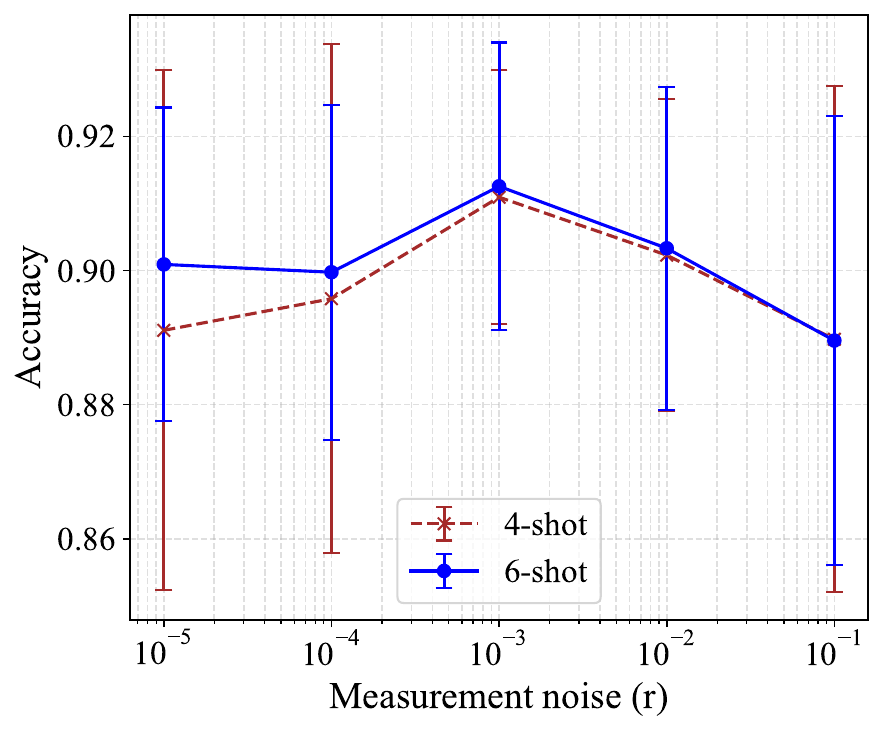}
    \caption{Accuracy vs. measurement noise for fixed test episodes and query per class.}
    \label{fig:measurement_noise}
\end{figure}

Figure~\ref{fig:measurement_noise} illustrates the effect of varying measurement noise \( r \) on classification accuracy under two support set sizes, with process noise fixed at \( q = 10^{-3} \). For the 4-shot setting, accuracy initially increases as \( r \) rises from \( 10^{-5} \) to \( 10^{-3} \), peaking at 91.09\% with \( r = 10^{-3} \). This trend suggests that mild measurement noise helps regularize the updates in the Kalman filter, smoothing over spurious variations in per-episode prototypes. However, further increasing \( r \) to \( 10^{-2} \) and \( 10^{-1} \) leads to a degradation in performance, falling back to 90.23\% and 88.97\% respectively. This decline implies that excessive measurement uncertainty downweights the contribution of new prototype observations, resulting in under-adaptive behavior. In the 6-shot case, a similar pattern is observed. Accuracy increases from 90.09\% at \( r = 10^{-5} \) to a peak of 91.25\% at \( r = 10^{-3} \), before declining as noise increases further. Notably, the 6-shot setting consistently outperforms 4-shot across all noise levels, reaffirming the value of additional support examples in stabilizing prototype estimation. The empirically optimal range for this dataset is centered around \( r = 10^{-3} \). In general, KPN demonstrates robustness to the choice of process noise within a reasonable range. However, these results highlight the importance of tuning the noise parameters for optimal performance of temporally regularized few-shot models.

\begin{figure}[!ht]
    \centering
    \includegraphics[width=\linewidth]{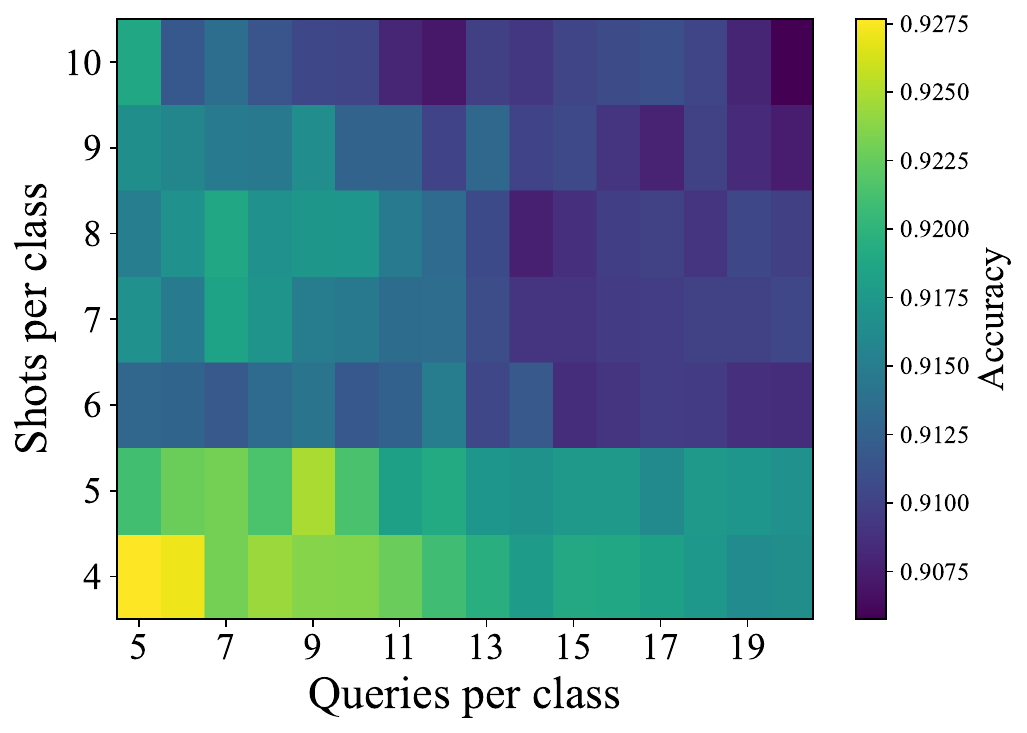}
    \caption{Average accuracy as a function of support size (shots) and query size per class.}
    \label{fig:shot_query_accuracy}
\end{figure}

To further evaluate the robustness of KPN, we performed a parameter sensitivity analysis with respect to the number of support examples (shots) and query examples per class. Specifically, we varied the number of shots from 4 to 10 and the query size per class from 5 to 20. The corresponding average classification accuracy across different configurations is visualized in Figure~\ref{fig:shot_query_accuracy}. The heatmap reveals several notable patterns in the effect of shot and query size. Increasing the number of support examples generally leads to improved classification accuracy. This trend aligns with common few-shot learning intuition, where larger support sets provide better prototype estimation. Interestingly, the highest performance is observed even with as few as 4 support examples, underscoring the effectiveness of the smoothed prototype mechanism in KPN.  The model also demonstrates consistent performance across a wide range of query sizes, with only slight variations in accuracy. Although a moderate increase in query size improves performance (due to more stable training gradients), the benefit diminishes beyond 10 queries per class, indicating that KPN is robust to changes in this parameter. In general, these results validate the stability and data efficiency of KPN in industrial fault detection tasks. The model remains performant across varying few-shot configurations, making it highly suitable for real-world diagnostic systems where labeled data may be limited.


\subsection{Test Episode Performance}
\begin{figure}[t!]\centering
    \includegraphics[width=\linewidth]{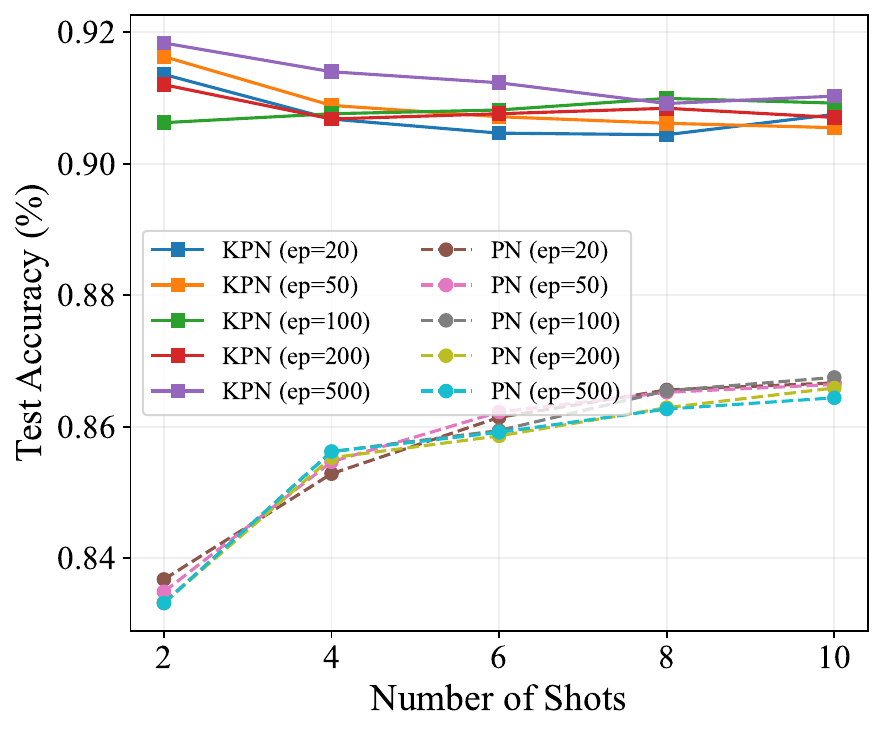}
    \caption{Accuracy vs. shots for different test episodes. }
    \label{fig:test_episodes_comparison}
\end{figure}

We further evaluate the performance of KPN under varying numbers of test episodes (50, 100, 200, and 500).
Figure \ref{fig:test_episodes_comparison} shows the few-shot classification accuracy as a function of the number of support samples per class (k-shot). For all test episode counts, KPN consistently outperforms PN across the entire range of shot values. These results further reinforce that Kalman smoothing of prototypes not only improves convergence during training but also leads to more reliable and stable generalization under various test-time conditions. This property is especially important for real-world gas turbine fault detection, where operational constraints may limit the number of available labeled or evaluation samples.

\subsection{Kalman Prototype Trajectory}

\begin{figure}[t!]\centering
    \includegraphics[width=\linewidth]{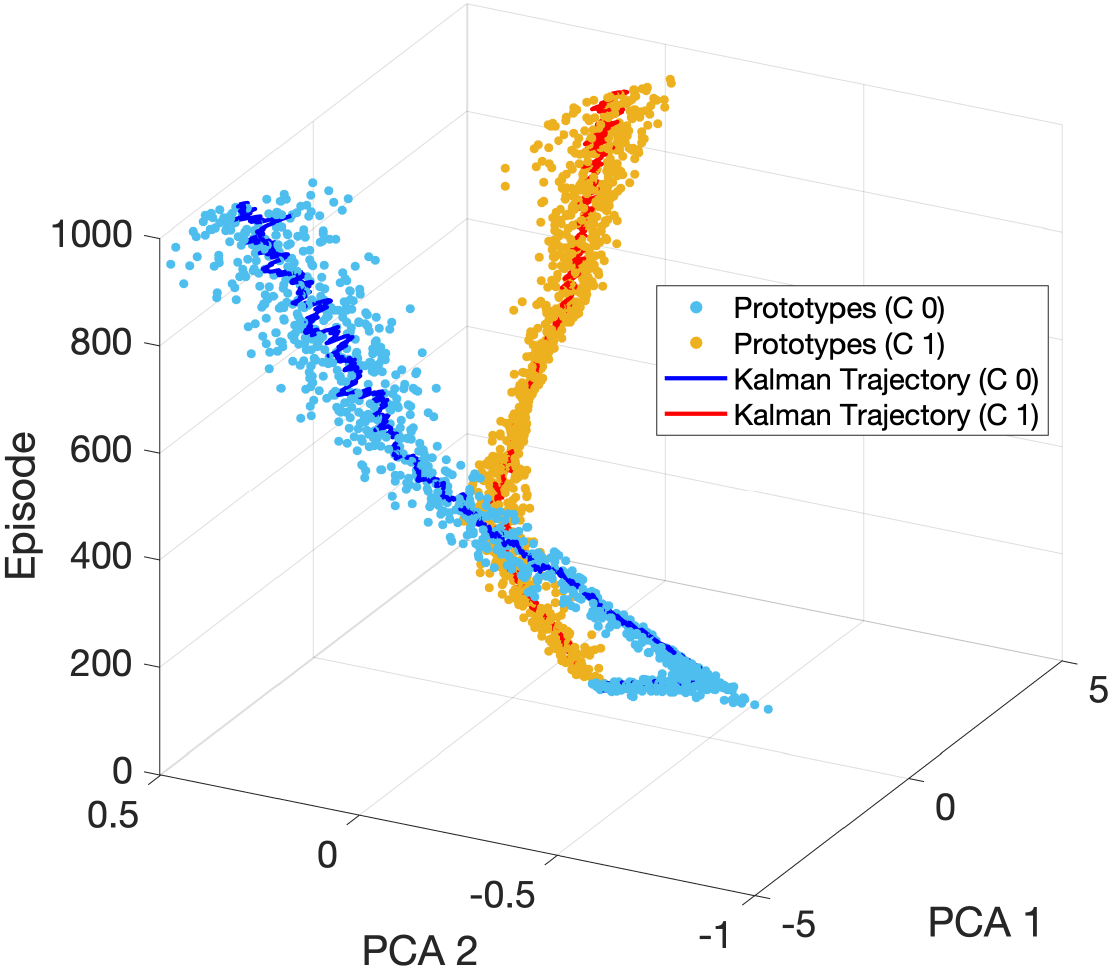}
    \caption{Prototypes Trajectory During Training. C0 and C1 denote classes with leak fault and  without leak fault, respectively.}
    \label{fig:kalman_4}
\end{figure}

We start our analysis by comparing the Kalman-filtered prototype with the noisy, scattered prototype standard trajectory during episodic training. Figure \ref{fig:kalman_4} presents the trajectories of class prototypes projected onto the first two principal components of the embedding space during 1000 episodic training. 
The observed prototype trajectories exhibit considerable variability and non-smoothness, indicative of episodic noise and high intra-class variance induced by few-shot support sampling. To mitigate the instability, a Kalman filter was applied independently to each class-specific prototype sequence (blue for Class 0 and red for Class 1). The filtering process utilized a process noise covariance $q=10^{-3}$
and observation noise covariance $r=10^{-2}$, corresponding to moderate model confidence in temporal smoothness relative to observation noise. The application of Kalman filtering yields two key improvements: the prototype trajectories become smoother, and the filtered trajectories maintain distinct separation between classes along the principal component axes, suggesting that discriminative structure in the latent space is enhanced. These results empirically validate the hypothesis that filtering reduces episodic variance and stabilizes class representations, enhancing robustness against stochasticity of few-shot training and improving fault classification boundaries.

\subsection{Training Convergence}

\begin{figure}[t!]\centering
    \includegraphics[width=\linewidth]{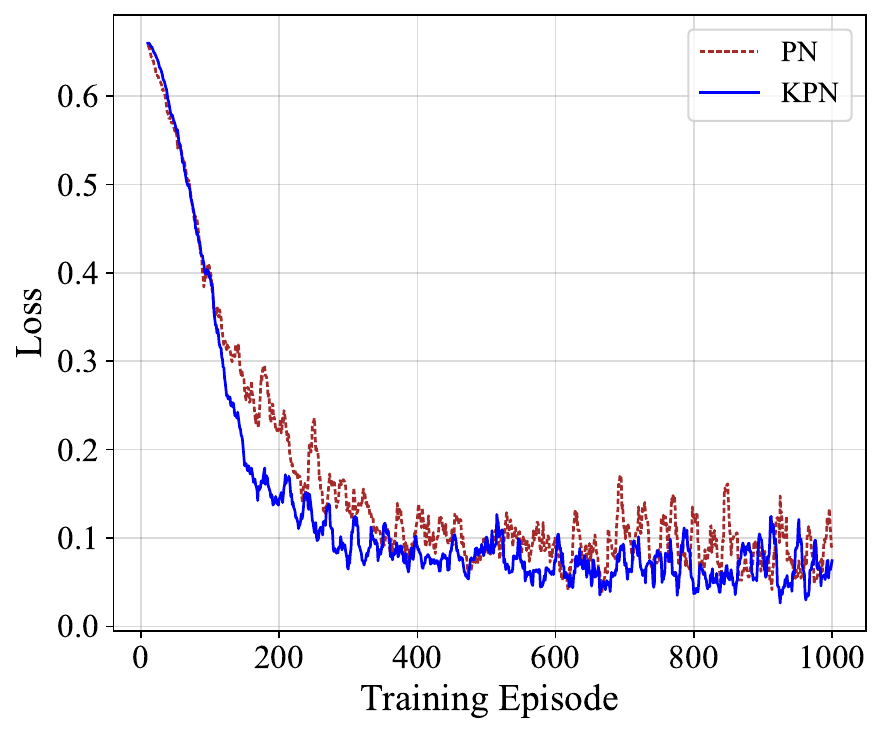}
    \caption{Training loss vs. iteration on the dataset. }
    \label{fig:loss_vs_episode_smooth}
\end{figure}

We train the proposed model using a Kalman-filtered prototype for each episode. Figure \ref{fig:loss_vs_episode_smooth} presents the training loss convergence of the standard prototypical network and the Kalman prototypical network over 1000 episodes. Both models are trained with identical few-shot settings. The PN curve exhibits significant oscillations throughout training, reflecting prototype instability due to episodic sampling noise. In contrast, KPN achieves smoother and more stable loss dynamics.
Early in training (episodes 0–300), KPN converges faster and maintains lower loss, indicating improved robustness against embedding drift. Overall, the training loss analysis confirms that integrating Kalman filtering into the prototypical network framework accelerates convergence in early training and reduces episodic variance—properties that are crucial for reliable few-shot learning in fault detection applications.

\section{Conclusion and Future Works}


In this work, we proposed the Kalman Prototypical Network (KPN) for few-shot fault detection in gas turbines. KPN is a robust prototypical network that integrates Kalman filtering to stabilize prototype estimation during episodic few-shot learning. Motivated by the observation that class prototypes evolve dynamically and noisily during training, we model prototype evolution as a latent stochastic process and apply temporal filtering to obtain denoised and temporally consistent prototypes. Through extensive experiments on a gas turbine fault detection dataset, we demonstrated that KPN consistently outperforms the standard Prototypical Network across a range of evaluation settings. Visualization of prototype trajectories revealed that KPN produces smoother and more stable class representations over training episodes. Training loss analysis showed that KPN reduces convergence noise and achieves faster and more stable optimization. Few-shot accuracy evaluations confirmed that KPN improves generalization across varying numbers of support shots, query sizes, and testing conditions, with particularly significant gains under low-shot and low-query regimes. Future work will explore adaptive filtering strategies, extend the approach to multiclass imbalanced settings, and investigate joint optimization of filter parameters alongside the embedding network.

\bibliographystyle{ieeetr}
\bibliography{references}

\end{document}